\documentclass[10pt,letterpaper]{article}
\usepackage{opex3}
\usepackage{hyperref}

\usepackage{graphicx,subfigure,epsfig,amsfonts,amsmath,amssymb,lscape,color,multirow}
\usepackage[noadjust]{cite}
\DeclareSymbolFont{grb}{OML}{cmm}{b}{it}
\DeclareMathSymbol{\alphab}{\mathord}{grb}{"0B}
\DeclareMathSymbol{\betab}{\mathord}{grb}{"0C}
\DeclareMathSymbol{\gammab}{\mathord}{grb}{"0D}
\DeclareMathSymbol{\deltab}{\mathord}{grb}{"0E}
\DeclareMathSymbol{\epsilonb}{\mathord}{grb}{"0F}
\DeclareMathSymbol{\zetab}{\mathord}{grb}{"10}
\DeclareMathSymbol{\etab}{\mathord}{grb}{"11}
\DeclareMathSymbol{\thetab}{\mathord}{grb}{"12}
\DeclareMathSymbol{\kappab}{\mathord}{grb}{"14}
\DeclareMathSymbol{\lambdab}{\mathord}{grb}{"15}
\DeclareMathSymbol{\mub}{\mathord}{grb}{"16}
\DeclareMathSymbol{\nub}{\mathord}{grb}{"17}
\DeclareMathSymbol{\rhob}{\mathord}{grb}{"1A}
\DeclareMathSymbol{\sigmab}{\mathord}{grb}{"1B}
\DeclareMathSymbol{\taub}{\mathord}{grb}{"1C}
\DeclareMathSymbol{\phib}{\mathord}{grb}{"1E}
\DeclareMathSymbol{\chib}{\mathord}{grb}{"1F}
\DeclareMathSymbol{\psib}{\mathord}{grb}{"20}
\DeclareMathSymbol{\omegab}{\mathord}{grb}{"21}
\DeclareMathSymbol{\epsilonb}{\mathord}{grb}{"22}
\DeclareMathSymbol{\varphib}{\mathord}{grb}{"27}

\begin{document}

\title{Coded aperture compressive\\ temporal imaging}

\author{Patrick Llull, Xuejun Liao, Xin Yuan, Jianbo Yang, David Kittle, Lawrence Carin, Guillermo Sapiro, and David J. Brady*}

\address{Fitzpatrick Institute for Photonics, Department of Electrical and Computer Engineering\\ Duke University, 129 Hudson Hall, Durham, North Carolina $27708$, USA}

\email{Corresponding author: dbrady@duke.edu} 



\begin{abstract}
We use mechanical translation of a coded aperture for code division multiple access compression of video.  We present experimental results for reconstruction at 148 frames per coded snapshot.
\end{abstract}


\bibliographystyle{plain}


\section{Introduction}
\label{sec:Intro}
Cameras capturing $>$ 1 gigapixel~\cite{brady2012multiscale}, $>$ 10,000 frames per second~\cite{kleinfelder200110000}, $>$ 10 ranges~\cite{ng2005light} and $>$ 100 color channels~\cite{brady2009optical} demonstrate that the optical data stream entering an aperture may approach 1 exapixel/second. While simultaneous capture of the all optical information is not precluded by diffraction or quantum limits, this capacity is beyond the capability of current electronics. Previous compressive sampling proposals to reduce read-out bandwidth paradoxically increase system volume\cite{brady2005compressive,shankar2010compressive} or operating bandwidth~\cite{takhar2006new,wakin2006compressive,hitomi2011video}. Here we use coded apertures to compress spatial and temporal sampling by $>$ 4x  and $>$ 10x, respectively, without substantially increasing system volume or power. We describe computational estimation of 148 high-speed temporal frames from a single captured frame. By combining physical layer compression as demonstrated here with sensor-layer compressive sampling strategies~\cite{zhang, oike2012256}, and by using multiscale design to parallelize read-out, one may imagine streaming full exapixel data cubes over practical communications channels.

Under ambient illumination, cameras commonly capture 10-1000 pW per pixel, corresponding to $10^9$ photons per second per pixel. At this flux, frame rates approaching $10^6$ per second may still provide useful information. Unfortunately, frame rate generally falls well below this limit due to  read-out electronics. The power necessary to operate an electronic focal plane is proportional to the rate at which pixels are read-out~\cite{iniewski2008circuits}. Current technology requires pW to nW per pixel, implying 1-100W per sq. mm of sensor area for full data optical data cube capture. This power requirement cascades as image data flows through pipeline of read-out, processing, storage, communications, and display. Given the need to maintain low focal plane temperatures, this power density is unsustainable.

While interesting features persist on all resolvable spatial and temporal scales, the true information content of natural fields is much less than the photon flux because diverse spatial, spectral and temporal channels contain highly correlated information~\cite{treeaporn2012space}.  Under these circumstances, feature specific~\cite{neifeld2003feature} and compressive~\cite{candes2008reflections,donoho2006compressed} multiplex measurement strategies developed over the past decade have been shown to maintain image information even when the number of digital measurement values is substantially less than the number of pixels resolved. Compressive measurement for visible imaging has been implemented using spatial light modulators (SLM) to code pixel values incident on a single detector~\cite{takhar2006new,wakin2006compressive}. Unfortunately, this strategy increases, rather than decreases, operating power and bandwidth because the increased data load in the encoding signal is much greater than the decreased data load of the encoded signal. If the camera estimates $F$ frames with $N$ pixels per frame from $M$ measurements, then $MN$ control signals enter the modulator to obtain $FN$ pixels. The bandwidth into the compressive camera exceeds the output bandwidth needed by a conventional camera by the factor $NC$, where $C$ is the compression ratio. Unless $C<1/N$, implying that the camera takes less than one measurement per frame, the control bandwidth exceeds the bandwidth necessary to fully read a conventional imager. This problem is slightly less severe in coding strategies derived from ``flutter shutter''~\cite{raskar2006coded} motion compensation strategies. The flutter shutter uses full frame temporal modulation to encode motion blur for inversion. Several studies have implemented per-pixel flutter shutter using spatial light modulators for video compression\cite{hitomi2011video, reddy2011p2c2, sankaranarayanan2012cs}. If we assume that these systems reconstruct at the full frame rate of the modulator, the control bandwidth is exactly equal to the bandwidth needed to read a conventional camera operating at the decompressed framerate. Alternatively, one may entirely avoid these problems by using parallel camera arrays with independent per pixel codes~\cite{brady2005compressive,shankar2010compressive} at the cost of increasing camera volume and cost by a factor of $M$.

Here we propose mechanical translation of a passive coded aperture for low power space-time compressive measurement. Coding is implemented by a chrome-on-glass binary transmission mask in an intermediate image plane. In contrast with previous approaches, modulation of the image data stream by harmonic oscillation of this mask requires no code transmission or operating power. We have previously used such masks for compressive imaging in coded aperture snapshot spectral imagers (CASSI)~\cite{gehm2007single}, which include an intermediate image plane before a spectrally dispersive relay optic. Here we demonstrate coded aperture compressive temporal imaging (CACTI). CASSI and CACTI share identical mathematical forward models. In CASSI, each plane in the spectral datacube is modulated by a shifted code. Dispersion through a grating or prism shifts spectral planes after coded aperture modulation. Detection integrates the spectral planes, but the datacube can be recovered by isloating each spectral plane based on its local code structure. This process may be viewed as code division multiple access (CDMA). In CACTI, translation of the coded aperture during exposure means that each temporal plane in the video stream is modulated by a shifted version of the code, thereby attaining per-pixel modulation using no additional sensor bandwidth.

Signal separation once again works by CDMA.  We isolate the object's temporal channels from the compressed data by inverting a highly-underdetermined system of equations.  By using an iterative reconstruction algorithm, we may estimate several high-speed video frames from a single coded measurement.

\section{Theory}
\label{sec:Theory}

\begin{figure}[hbtp]
\centering
\includegraphics[width=3.25in]{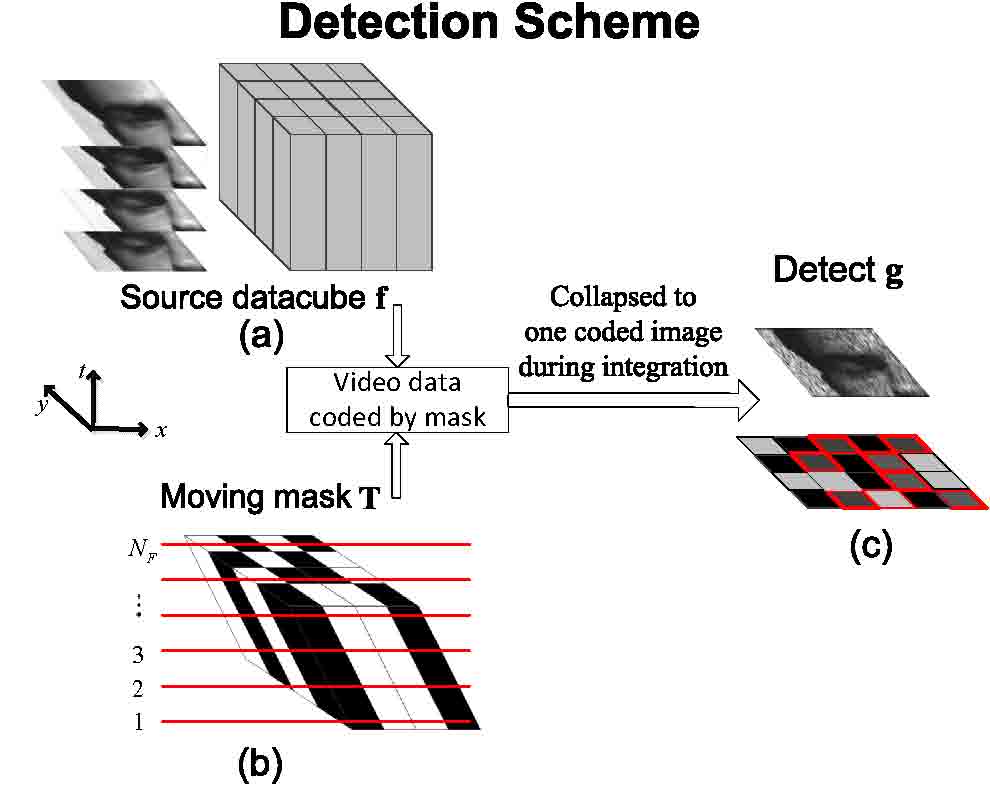}
\caption{Detection process.  (a) A discrete space-time source datacube is (b) multiplied at each of $N_F$ temporal channels with a shifted version of a coded aperture pattern.
(c) Each detected frame ${\bf g}$ is the summation of the coded temporal channels and contains the object's spatiotemporal-multiplexed information. The dark grey (red-outlined) and black detected pixels in (c) pictorially depict the code's location at the beginning and the end of the camera's integration window, respectively.}
\label{detectionfig}
\end{figure}

One may view CACTI's CDMA sensing process as uniquely patterning high-speed spatiotemporal object voxels $f(x,y,t)\in {\mathbb R^3}$ with a transmission function that shifts in time (Fig. \ref{detectionfig}).  Doing this applies distinct local coding structures to each temporal channel prior to integrating the channels as limited-framerate images $g(x',y',t') \in {\mathbb R^2}$ on the $N$-pixel detector.  An $N_F$-frame, high-speed estimate of $f(x,y,t)$ may be reconstructed from each low-speed coded snapshot $g(x',y',t'), {\rm with}~t' < t$.

Considering only one spatial dimension ($(x,y) \rightarrow x$) and respectively denoting object-and image-space coordinates with unprimed and primed variables, the sampled data $g(x',t')$ consists of discrete samples of the continuous transformation~\cite{brady2009optical}

\begin{equation}
g(x',t') = \int_1^{N_F}\int_1^{N} f(x,t)T(x-s(t))
{\rm rect}\left(\frac{x-x'}{\Delta_x}\right){\rm rect}\left(\frac{t-t'}{\Delta_t}\right)
dxdt,
\label{eqn:theBigEquation}
\end{equation}

\noindent where $T(x-s(t))$ represents the transmission function of the coded aperture,
$\Delta_x$ is the detector pixel size, ${\rm rect}(\frac{x}{\Delta_x})$ is the pixel sampling function and $\Delta_t$ is the temporal integration time.  $s(t)$ describes the coded aperture's spatial position during the camera's integration window.

One may analyze the expected temporal resolution of the coded data by considering the Fourier transform of Eq.~(\ref{eqn:theBigEquation}).
Assuming the coded aperture moves linearly during $\Delta_t$ such that $s(t)=\nu t$, the image's temporal spectrum is given by

\begin{equation}
{\hat g}(u,v) = {\rm sinc}(u\Delta_x ){\rm sinc }(v\Delta_t )\int {\hat f}(u-w,v-\nu w){\hat T}(w)dw,
\label{eqn:FTBigEquation}
\end{equation}

\noindent where ${\hat f}(u,v)$ is the 2D Fourier transform of the space-time datacube and ${{\hat T}}(w)$ is the 1D Fourier transform of the spatial code.
Without the use of the coded aperture, ${\hat g}(u,v)= {\rm sinc}(u\Delta_x){\rm sinc }(v\Delta_t){\hat f}(u,v)$ and the sampled data stream is
proportional to the object video low-pass filtered by the pixel sampling functions. Achievable resolution is proportional to $\Delta_x$ in $x$ and $\Delta_t$ in time. The moving code aliases higher frequency components of the object video into the passband of the detector sampling functions.
The support of ${{\hat T}}(w)$ extends to some multiple of the code feature size $\Delta_c$ (in units of detector pixels), meaning that the effective passband may be increased by a factor proportional to $1/\Delta_c$ in space and $\nu/\Delta_c$ in time.  In practice, finite mechanical deceleration times cause ${\hat  T}(w)$ to have significant DC and low-frequency components in addition to the dominant $\nu = \frac{C}{\Delta_t}$; hence, high and low frequencies alike are aliased into the system's passband.

%

\begin{figure}[hbtp]
\centering
\includegraphics[width=10cm]{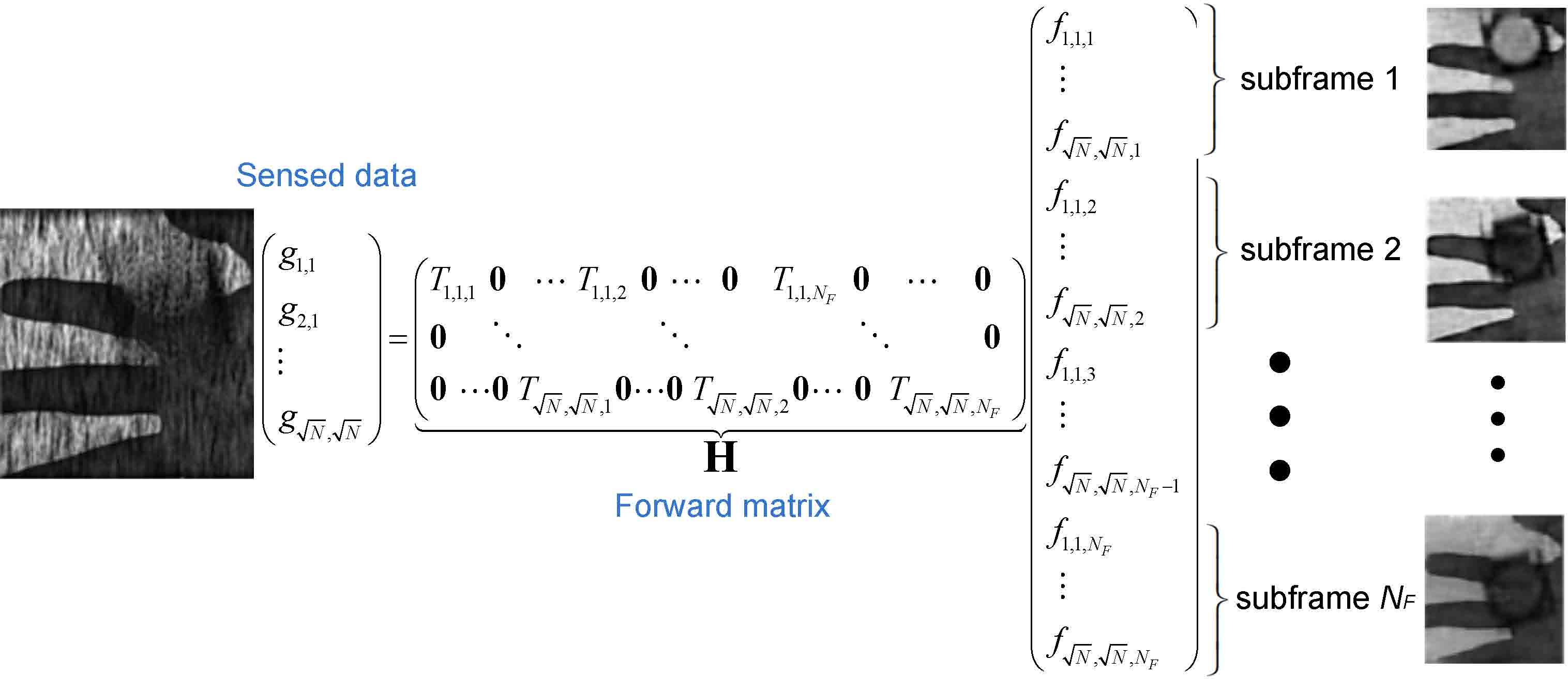}
\caption{Linear system model.  $N_F$ subframes of high-speed data ${\bf f}$ are estimated from a single snapshot ${\bf g}$.  The forward model matrix ${\bf H}$ has many more columns than rows and has dimensions $N\times (N\times N_F)$.}
\label{detectionfigandlayout}
\end{figure}

Considering a square $N$-pixel active sensing area, the discretized form of the three-dimensional scene is ${\bf f} \in {\mathbb R}^{\sqrt{N}\times \sqrt{N} \times N_F}$, i.e. a $({\sqrt{N}\times \sqrt{N}})\times N_F$-voxel spatiotemporal datacube.  In CACTI, a time-varying spatial transmission pattern ${\bf T} \in {\mathbb R}^{\sqrt{N}\times \sqrt{N} \times N_F}$ uniquely codes each of the $N_F$ temporal channels of ${\bf f}$ prior to integrating them into one detector image ${\bf g}\in {\mathbb R}^{\sqrt{N}\times \sqrt{N} }$ during $\Delta_t$.
These measurements at spatial indices $(i,j)$ and temporal index $k$ are given by


\begin{eqnarray}
g_{i,j} &=& \sum_{k=1}^{N_F}T_{i,j,k} f_{i,j,k} + n_{i,j},
\label{eqn:linearProcess}
\end{eqnarray}
%

\noindent where $n_{i,j}$ represents imaging noise at the $(i,j)^{th}$ pixel.  One may rasterize the discrete object ${\bf f} \in {\mathbb R}^{NN_F\times 1}$, image ${\bf g} \in {\mathbb R}^{N\times 1}$, and noise ${\bf n} \in {\mathbb R}^{N\times 1}$ to obtain the linear transformation given by
\begin{eqnarray}
{\bf g} &=& {\bf H}{\bf f} + {\bf n},
\label{eqn:linearProcess}
\end{eqnarray}

\noindent where ${\bf H} \in {\mathbb R}^{N\times NN_F}$ is the system's \textit{discrete forward matrix} that accounts for sampling factors including the optical impulse response, pixel sampling function, and time-varying transmission function.  The forward matrix is a 2-dimensional representation of the 3-dimensional transmission function ${\bf T}$:
\begin{eqnarray}
{\bf H}_k &\stackrel{\rm def}{=}& {\rm diag}\left[T_{1,1,k}~~T_{2,1,k}~~\cdots~~T_{\sqrt{N},\sqrt{N},k}\right],~~~k=1,\dots,N_F;\\
{\bf H} &\stackrel{\rm def}{=} & \left[{\bf H}_1 ~~ {\bf H}_2~~ \cdots ~~{\bf H}_{N_F}\right],
\end{eqnarray}
\noindent where ${\bf H}_k \in {\mathbb R}^{N\times N}$ is a matrix containing the entries of ${\bf T}_k$ along its diagonal and ${\bf H}$ is a concatenation of all ${\bf H}_k$, $k \in{\{1,\dots,N_F\}}$.  Fig. \ref{detectionfigandlayout} underlines the role {\bf H} plays in the linear transformation.



%


\begin{figure}[hbtp]
\centering
\includegraphics[width=3.25in]{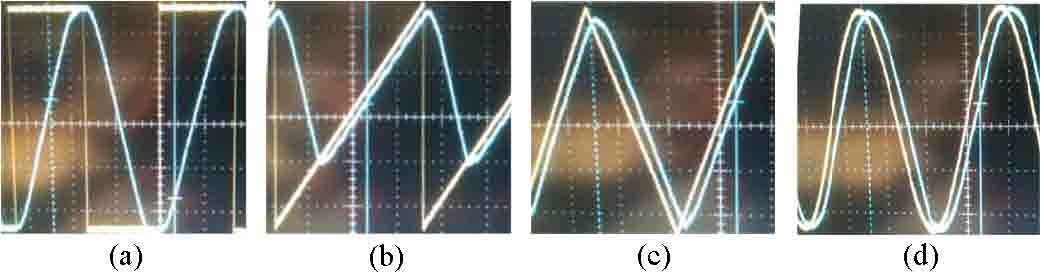}
\caption{Waveform choices for $s(t)$.  Yellow: signal from function generator.  Blue: actual hardware motion. Note the poor mechanical response to sharp rising/falling edges in (a) and (b).  The sine wave (d) is unpreferable because of the nonuniform exposure time of different ${\bf T}_k$.}
\label{fig:waveforms}
\end{figure}

\begin{figure}[hbtp]
\centering
\includegraphics[width=3.25in]{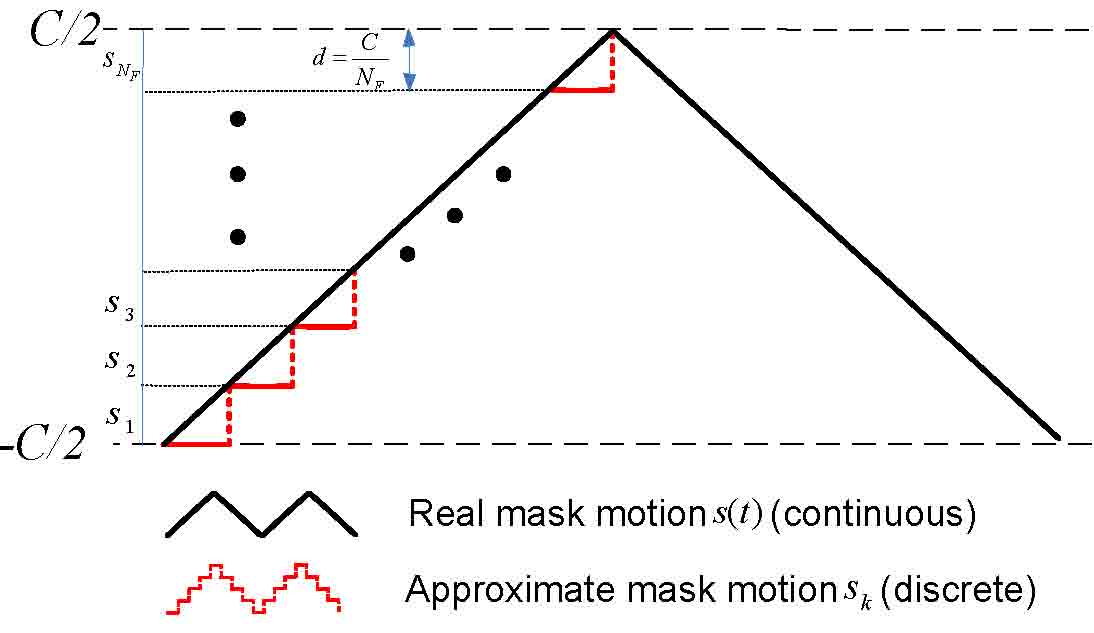}
\caption{Continuous motion and discrete approximation to coded aperture movement during integration time.  The discrete triangle function $s_k$ more-accurately approximates the continuous triangle wave driving the mask with smaller values of $d$ but adds more columns to {\bf H}.}
\label{fig:motion}
\end{figure}

At the $k^{th}$ temporal channel, the coded aperture's transmission function ${\bf T}$ is given by
\begin{equation}
{\bf T}_{k} =  {\rm Rand}(\sqrt{N},\sqrt{N},s_k),
\label{eqn:transmissionfunction}
\end{equation}
\noindent where ${\rm Rand}(m,n,p)$ denotes a $50\%$, $m\times n$ random binary matrix shifted vertically by $p$ pixels (optimal designs could be considered for this system as well).  $s_k$ discretely approximates $s(t)$ at the $k^{th}$ temporal channel by 

\begin{equation}
s_k = C{\rm Tri}\left[\frac{k}{2\Delta_t}\right]
\label{eqn:position}
\end{equation}

\noindent where $C$, the system's compression ratio, is the amplitude traversed by the code in units of detector pixels.  ${\rm Tri}\left[\frac{k}{2\Delta_t}\right]$ represents a discrete triangle wave signal of twice the integration time periodicity.  The camera integrates during the $C$-pixel sweep of the coded aperture on the image plane and detects a linear combination of $C$ uniquely-coded temporal channels of $f$.
%

\begin{figure}[hbtp]
\centering
\includegraphics[width=3.25in]{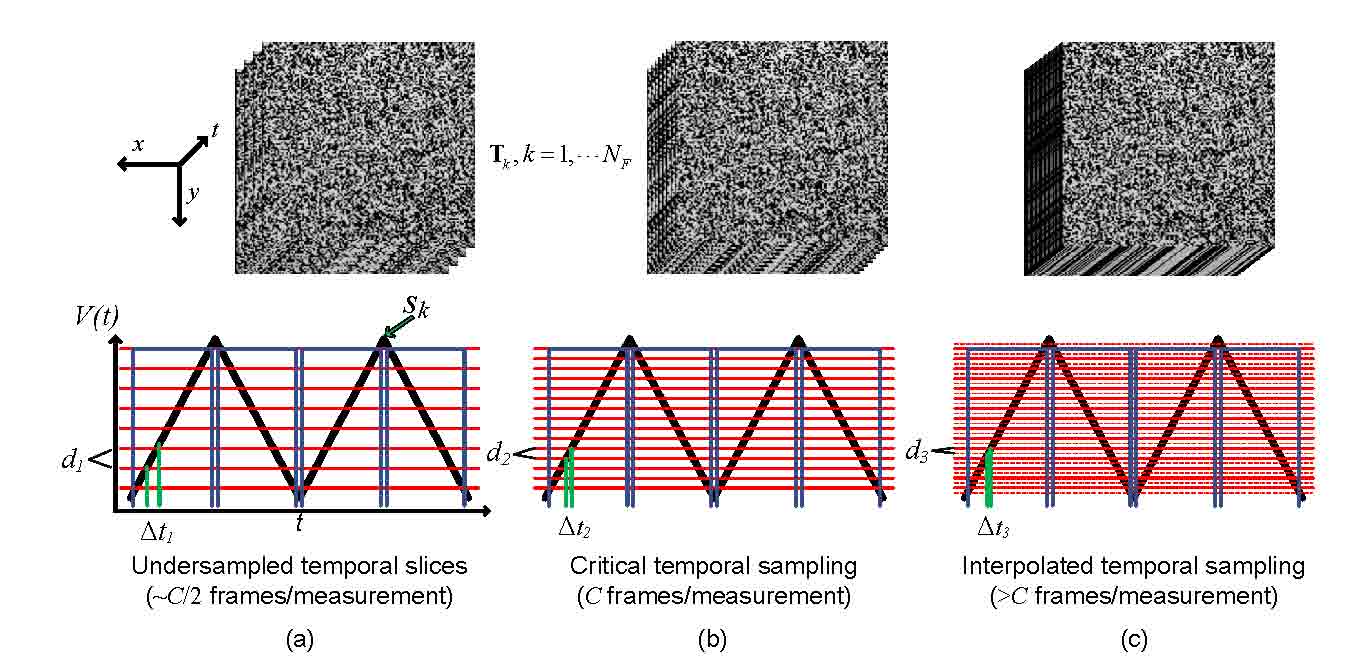}
\caption{Temporal channels used for the reconstruction.  Red lines indicate which subset of transverse mask positions $s_k$ were utilized to construct the forward matrix {\bf H}.  Blue lines represent the camera integration windows.  (a) Calibrating with fewer $s_k$ results in a better-posed inverse problem but doesn't as closely approximate the temporal motion $s(t)$. (b) With $d = 1$, each pixel integrates several unique coding patterns with a temporal separation $\Delta t = C^{-1}$. (c) Constructing $\bf H$ with large $N_F$ ($d < 1$) interpolates the motion occurring between the uniquely-coded image frames.}
\label{fig:temporalsamplingcalcubes}
\end{figure}

Periodic triangular motion lets us use the same discrete forward matrix ${\bf H}$ to reconstruct any given snapshot ${\bf g}$ within the acquired video while adhering to the hardware's mechanical acceleration limitations (Fig. \ref{fig:waveforms}). The discrete motion $s_k$ (Eq. (\ref{eqn:position})) closely approximates the analog triangle waveform supplied by the function generator (Fig. \ref{fig:motion}).

Let $d$ represent the number of \textit{detector} pixels the mask moves between adjacent temporal channels $s_k$ and $s_{k+1}$.  $N_F$ frames are reconstructed from a single coded snapshot given by

\begin{equation}
N_F = \frac{C}{d},
\label{eqn:numFrames}
\end{equation}


\noindent thus, altering $d$ will affect the number of reconstructed frames for given a compression ratio $C$.  

In the case of $d = 1$ (i.e. $N_F = C$), the detector pixels that sense the continuous, temporally-modulated object $f$ are {\em critically-encoded}; each pixel integrates a series of nondegenerate mask patterns (Fig. \ref{fig:temporalsamplingcalcubes}(b)) during $\Delta_t$.

When $d < 1$ (i.e. $N_F > C$), every $\frac{1}{d}$ temporal channels of ${\bf H}$ will contain nondegenerate temporal code information.  These channels will reconstruct as if the sensing pixels are critically-encoded.  The other temporal slices will interpolate the motion {\em between} critically-encoded temporal channels (Fig. \ref{fig:temporalsamplingcalcubes}(c)).  Generally, this interpolation accurately estimates the direction of the motion between these critically-encoded states but retains most of the residual motion blur.

\section{Experimental Hardware}
\label{Sec:Hardware}
\begin{figure}[hbtp]
\centering
\includegraphics[width=3.25in]{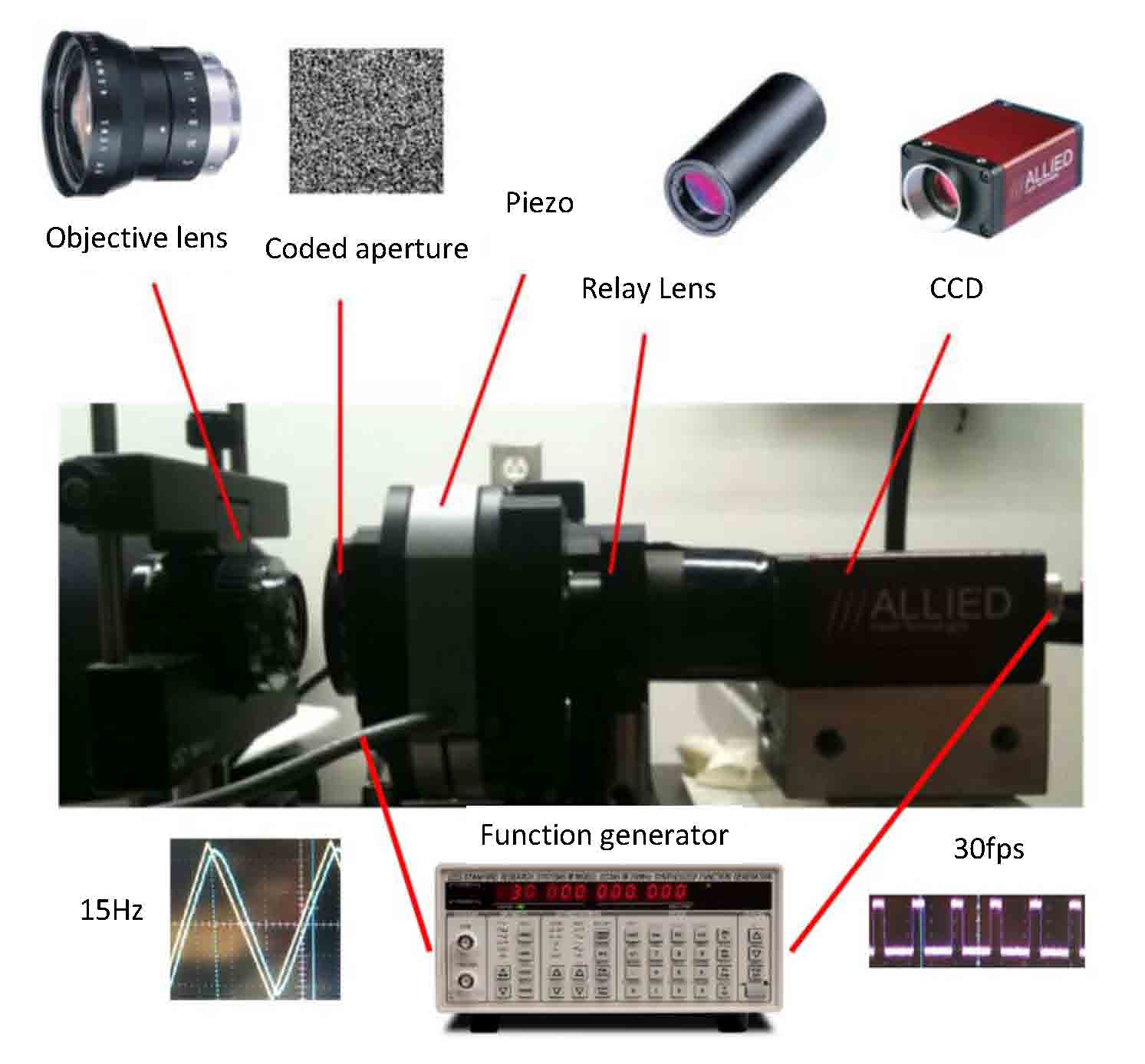}
\caption{CACTI Prototype hardware setup.  The coded aperture is $5.06\rm{mm} \times 4.91\rm{mm}$ and spans 248 $\times$ 256 detector pixels.  The function generator moves the coded aperture and triggers camera acquisition with signals from its function and SYNC outputs, respectively.}
\label{fig:Hardware}
\end{figure}

The experimental prototype camera (Fig. \ref{fig:Hardware}) consists of a 50mm camera objective lens (Computar), a lithographically-patterned chrome-on-quartz coded aperture with anti-reflective coating for visible wavelengths~\cite{gehm2007single} (Eq. (\ref{eqn:transmissionfunction})) mounted upon a piezoelectric stage (Newport Co.), an $F/8$ achromatic relay lens (Edmund Optics), and a $\rm{640}\times \rm{480}$ FireWire IEEE 1394a monochrome CCD camera (Marlin AVT).  

The objective lens images the continuous scene $f$ onto the piezo-positioned mask.  The function generator (Stanford Research Systems DS345) drives the piezo with a 10V pk-pk, 15Hz triangle wave to locally code the image plane while the camera integrates.  We operate at this low frequency to accommodate the piezo's finite mechanical deceleration time.

To ensure the CDMA process remains time-invariant, we use the function generator's SYNC output to generate a 15Hz square wave.  We frequency-double this signal using an FPGA device (Altera) to trigger camera integrations once along the mask's upward and downward motion (Fig. \ref{fig:temporalsamplingcalcubes}).

The relay lens images the spatiotemporally modulated scene onto the camera, which saves the 30fps coded snapshots to a local computer.  $N_F$ video frames of the discrete scene ${\bf f}$ are later reconstructed from each coded image ${\bf g}$ offline by the Generalized Alternating Projection (GAP)~\cite{liao2012GAP} algorithm.

During $\Delta_t$, the piezo can move a range of $0-160\mu m$ vertically in the $(x,y)$ plane.  Using $158.4\mu m$ of this stroke moves the coded aperture eight $19.8\mu m$ elements (sixteen $9.9\mu m$ detector pixels) during each camera integration period $\Delta_t$.  Using larger strokes for a given modulation frequency is possible and would increase $C$.

\begin{figure}[hbtp]
\centering
\includegraphics[width=3.25in]{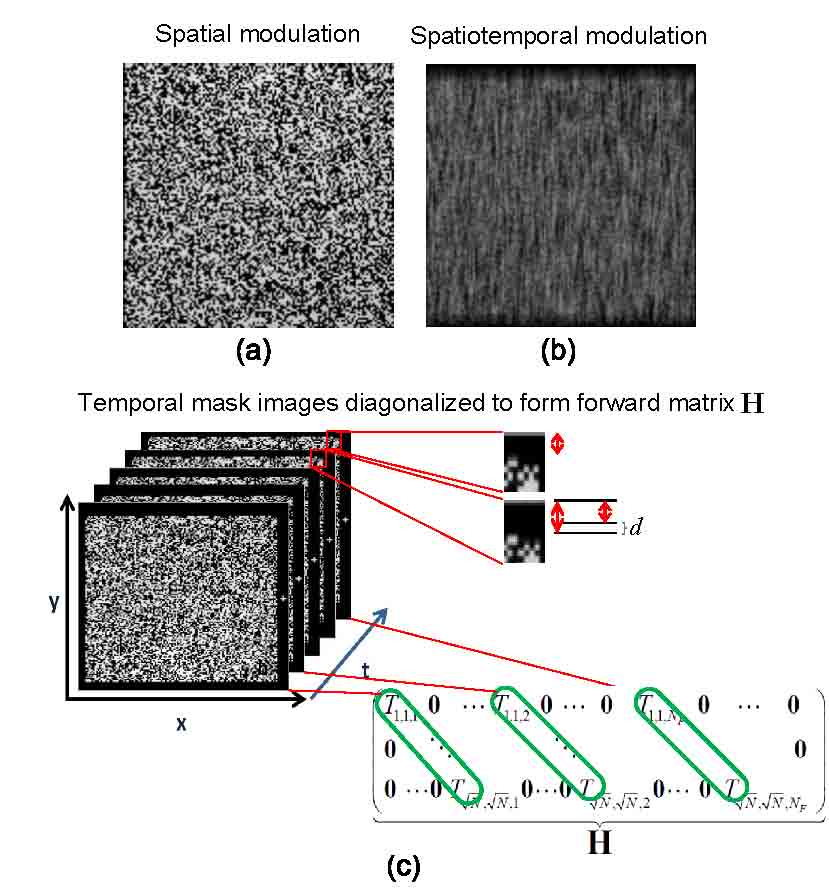}
\caption{Spatial and temporal modulation.  (a) A stationary coded aperture spatially modulates the image.  (b) Moving the coded aperture during the integration window applies local code structures to $N_F$ temporal channels, effectively shearing the coded space-time datacube and providing per-pixel flutter shutter.  (c) Imaging the (stationary) mask at positions $d$ pixels apart and storing them into the forward matrix ${\bf H}$ simulates the mask's motion, thereby conditioning the inversion.}
\label{spatiotemporalmodulation}
\end{figure}

Importantly, using a piezoelectric stage itself is not the optimal solution to translate a coded aperture during $\Delta_t$.  This device was preferable for the hardware prototype because of its precision and convenient built-in Matlab interface.  However, a low-resistance spring system could, in principle, serve the same purpose while using very little power.

\subsection{Forward Model Calibration} \label{sec:calibration}

%

We calibrate the system forward model by imaging the aperture code under uniform, white illumination at discrete spatial steps $s_k$ according to Eq. (\ref{eqn:position}).  Steps are $d$ detector pixels apart over the coded aperture's range of motion (Figs. ~\ref{spatiotemporalmodulation}(c),\ref{fig:temporalsamplingcalcubes},).  This accounts for system misalignments and relay-side aberrations.  A Matlab routine controls the piezoelectric stage position during calibration.  Since Matlab cannot generate a near-analog waveform for continuous motion, we connect the piezoelectric motion controller to the function generator via serial port during experimental capture.  

We use an active area of $281\times 281$ detector pixels to account for the $248\times 256$-pixel coded aperture's motion $s_k$ with additional zero-padding.  We choose $d = \frac{\Delta_x}{10}$ to provide a substantial basis with which to construct the forward model while remaining well above the piezo's $0.0048$ pixel RMS jitter.  Storing every temporal channel spaced $d = 0.99\mu m$ apart into {\bf H} results in $N_F = 160$ reconstructed frames.



When reconstructing, we may diagonalize any subset of temporal slices of the $281\times281\times160$ set of mask images into the forward model (Fig. \ref{spatiotemporalmodulation}).  We found the optimum subset of mask positions within this 160-frame set of $s_k$ through iterative $||{\bf g}-{\bf Hf_e}||_2$-error reconstruction tests, where ${\bf f_e}$ is GAP's $N_F$-frame estimate of the continuous motion $f$.  From these tests, we chose and compared two numbers of frames to reconstruct per measurement, $N_F = C = 14$ and $N_F = 148$.  {\bf H} has dimensions $281^2\times(281^2\times N_F)$ for both of these cases.

\begin{figure}[hbtp]
\centering
\includegraphics[width=3.25in]{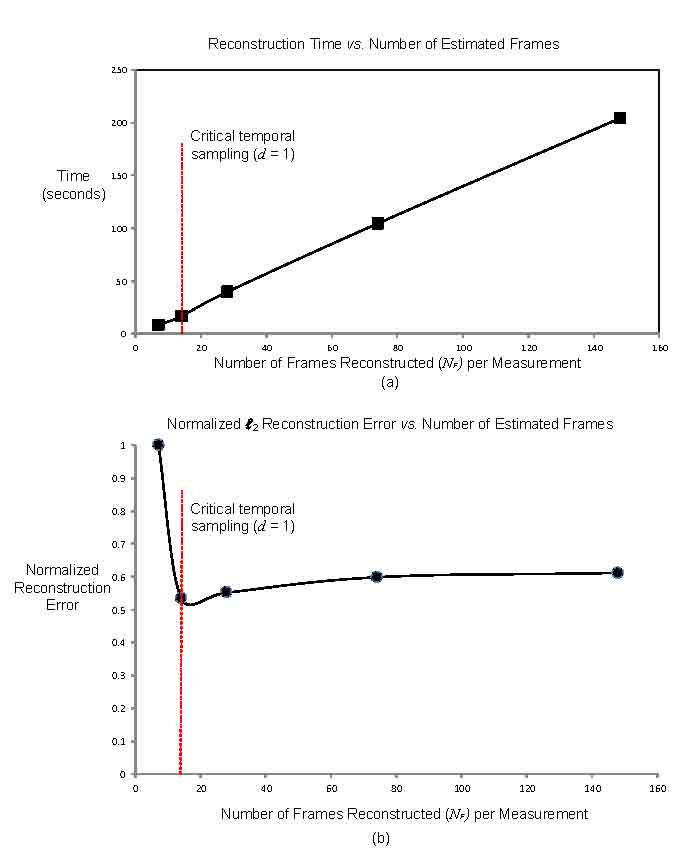}
\caption{Algorithm convergence times and relative residual reconstruction errors for various compression ratios.  (a) GAP's reconstruction time increases linearly with data size.  Tests were performed on a ASUS U46E laptop (Intel quad core I7 operated at 3.1GHz. (b) Normalized $\ell_2$ reconstruction error vs. number of reconstructed frames.  The residual error reaches a minimum at critical temporal sampling and gradually flattens out with finer temporal interpolation (lower $d$).}
\label{reconstructiontime}
\end{figure}

As seen in Figs. \ref{face}-\ref{water}, decreasing $d$ and estimating up to 148 frames from a single exposure ${\bf g}$ does not significantly reduce the aesthetic quality of the inversion results, nor does it significantly affect the residual error (Fig. \ref{reconstructiontime}(b)).  The reconstruction time increases approximately linearly with $N_F$ as shown in Fig. \ref{reconstructiontime}(a).

\section{Reconstruction Algorithm}
\label{Sec:Algorithm}

\begin{figure}[hbtp]
\centering
\includegraphics[width=3.25in]{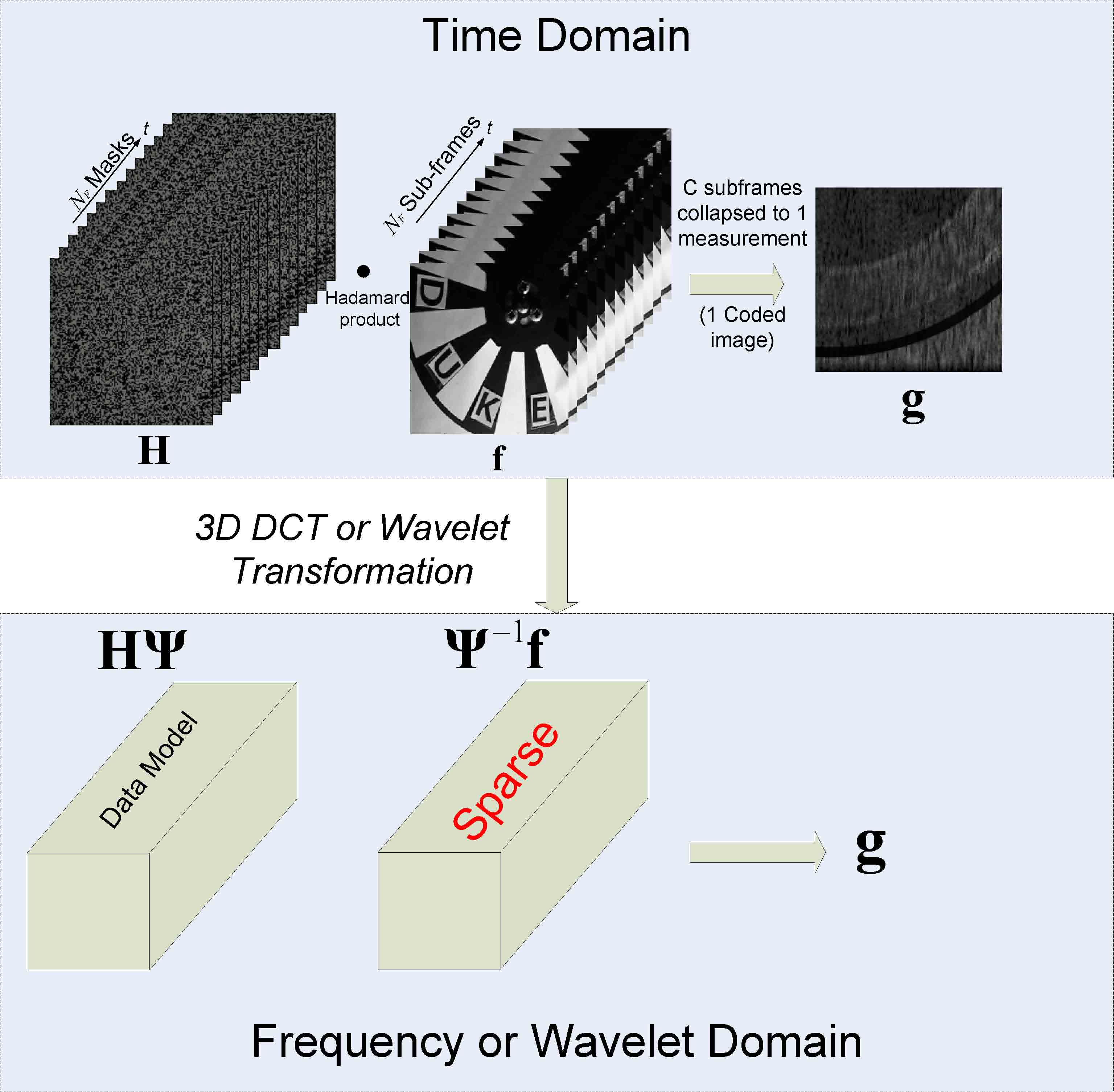}
\caption{Illustration of the GAP algorithm.}
\label{Fig_GAPforPaper}
\end{figure}
Since ${\bf H}$ multiplexes many local code patterns of the continuous object to the discrete-time image ${\bf g}$, inverting Eq. (\ref{eqn:linearProcess}) for ${\bf f}$ becomes difficult as $N_F$ increases.  Least-squares, pseudoinverse, and other linear inversion methods cannot accurately reconstruct such underdetermined systems.  We use an iterative reconstruction algorithm called Generalized Alternating Projection (GAP) that exploits image and video models (priors) to effectively solve this ill-posed inverse problem \cite{liao2012GAP}.

GAP takes advantage of the structural sparsity of the subframes in transform domains such as wavelets and discrete cosine transform (DCT). Fig. \ref{Fig_GAPforPaper} illustrates the underlying principle of GAP.  It is worth noting that GAP is based on the volume's global sparsity and requires no training whatsoever.  In other words, GAP is a universal
reconstruction algorithm insensitive to data being inverted.

The GAP algorithm makes use of Euclidean projections on two convex sets, which respectively enforce data fidelity and structural sparsity.
Please refer to \cite{liao2012GAP} for details.
Furthermore, GAP is an anytime algorithm; the results produced by the algorithm converge monotonically to the true value as the computation proceeds.  The monotonicity has been generally observed in our extensive experiments and theoretically established under a set of sufficient conditions on the forward model.  The reconstructed subframes continually improve over successive iterations and the user can halt computation at anytime to obtain intermediate results.  The user may then continue improving reconstruction by resuming the computation.  The following briefly reviews the main steps of the GAP algorithm \cite{liao2012GAP}.

\subsection{The Linear Manifold}
Data fidelity is ensured by projection onto the linear manifold $\Pi={\bf f}: \{\sum^{N_F}_{k=1}{ H}_{i,j,k}{f}_{i,j,k} + {n}_{i,j}={g}_{i,j},\,\,\forall\,\,(i,j)\}$, which consists of all legitimate high-speed frames of ${\bf f}$ integrated onto the detector via Eq. (\ref{eqn:linearProcess}).  In other words, $\Pi$ is a set of solutions to the underdetermined system of linear equations which are disambiguated by using structural sparsity.

\subsection{The Weighted $\ell_{2,1}$ Ball}
Structural sparsity is encoded by $G=\{G_{1}, G_{2}, \cdots, G_{m}\}$, a partition of the indices $\{(i,j,k)$ that span the voxels of ${\bf f}$, and the associated weights $\beta=\{\beta_{l}:\beta_{l}>0,l=1,2,\cdots,m\}$. The weighted $\ell_{2,1}$ ball is defined as $\Lambda(R)=\left\{{\bf f}:\|Q({\bf f})\|_{\ell_{2,1}^{G\beta}}\leq{R}\right\}$, where $Q({\bf f})$ is an orthonormal transform (wavelet transformation or Discrete Cosine Transformation (DCT)) of the volume ${\bf f}$,
\begin{eqnarray}
Q({\bf f})_{i,j,k}&=& \sum_{i',j',k'}Q_{1}(i,i')Q_{2}(j,j')Q_{3}(k,k'){f}_{i',j',k'},
\nonumber
\end{eqnarray}
with $Q_{1}$,$Q_{2}$,$Q_{3}$ orthonormal matrices, and
\begin{eqnarray}\label{eq:w-L21-norm-define}
\|\bf {w}\|_{\ell_{2,1}^{G\beta}}&=&\sum_{l=1}^{m}\beta_{l}\sqrt{\sum_{(i,j,k)\in{}G_{l}}{w}_{i,j,k}^{2}}
\nonumber
\end{eqnarray}
is a weighted $\ell_{2,1}$ norm of ${\bf w}$. Note that $\Lambda(R)$ is constructed as a weighted $\ell_{2,1}$ ball in the space of transform coefficients ${\bf w}=Q({\bf f})$, since structural sparsity is desired for the coefficients instead of voxels. The ball is rotated in the voxel space due to the orthonormal transform $Q(\cdot)$.

\subsection{Euclidean Projections}
The Euclidean projection of $\theta$ onto $\Pi$,
\begin{eqnarray}
P_{\Pi}(\theta)=\mathrm{arg}\min_{{f}:{f}\in\Pi}\sum_{i,j,k}({\theta}_{i,j,k}-{f}_{i,j,k})^{2},
\nonumber
\end{eqnarray}
is given by
\begin{eqnarray}\label{eq:proj-linear}
{f}_{i,j,k}&=& {\theta}_{i,j,k}+\frac{{H}_{i,j,k}}{\sum_{k'=1}^{N}{H}_{i,j,k'}^{2}}\left({g}_{i,j}-\sum_{k'=1}^{N}{\theta}_{i,j,k'}{H}_{i,j,k'}\right).
\end{eqnarray}
The Euclidean projection of ${\bf f}$ onto $\Lambda(R)$,
\begin{eqnarray}
P_{\Lambda(R)}({\bf f})=\mathrm{arg}\min_{{\theta}:{\theta}\in\Lambda(R)}\sum_{i,j,k}({\theta}_{i,j,k}-{f}_{i,j,k})^{2},
\nonumber
\end{eqnarray}
can be equivalently written as
\begin{eqnarray}
P_{\Lambda(R)}({\bf f})=Q^{-1}\left(\mathrm{arg}\min_{{\boldsymbol\theta}:\|{\boldsymbol\theta}\|_{\ell_{2,1}^{G\beta}}\leq{}R}\sum_{i,j,k}({\boldsymbol\theta}_{i,j,k}-Q({\bf f})_{i,j,k})^{2}\right).
\nonumber
\end{eqnarray}
using that fact that Euclidean distance is invariant to the orthonormal transform $Q(\cdot)$.  We are only interested in $P_{\Lambda(R)}({\bf f})$ when $R$ takes the special values considered below.

\subsection{Alternating Projection between $\Pi$ and $\Lambda(R)$ when $R$ Systematically Changes}
The GAP algorithm is a sequence of Euclidean projections between a linear manifold and a weighted $\ell_{2,1}$ ball that undergoes a systematic change in size. Starting from ${\boldsymbol\theta}^{(0)}=0$, the GAP algorithm iterates between the following two steps, until $\|{\bf f}^{(t)}-{\boldsymbol\theta}^{(t)}\|_{\ell_{2}}$ converges in $t$.

\begin{enumerate}
\item[1)] Projection on the linear manifold,
\begin{eqnarray}
{\bf f}^{(t)}&=&P_{\Pi}({\boldsymbol\theta}^{(t-1)}),\quad{}t\geq1,
\nonumber
\end{eqnarray}
with the solution given in Eq. (\ref{eq:proj-linear}).
\item[2)] Projection on the weighted $\ell_{2,1}$ ball of changing size,
\begin{eqnarray}
{\boldsymbol\theta}^{(t)}&=&P_{\Lambda(R^{(t)})}({\bf f}^{(t)}),\quad{}t\geq1,
\nonumber
\end{eqnarray}
where
\begin{eqnarray}\label{eq:w-L21-norm-define}
R^{(t)}&=&\sum_{q=1}^{m^{\star}}\beta_{l_{q}}^{2}\left(\frac{\sqrt{\sum_{(i,j,k)\in{}G_{l_{q}}}Q({\bf f}^{(t)})_{i,j,k}^{2}}}{\beta_{l_{q}}}-\frac{\sqrt{\sum_{(i,j,k)\in{}G_{l_{m^{\star}+1}}}Q({\bf f}^{(t)})_{i,j,k}^{2}}}{\beta_{l_{m^{\star}+1}}}\right),
\nonumber
\end{eqnarray}
$(l_{1},l_{2},\cdots,l_{m})$ is a permutation of $(1,2,\cdots,m)$ such that
\begin{eqnarray}
\frac{\sqrt{\sum_{(i,j,k)\in{}G_{l_{q}}}Q({\bf f}^{(t)})_{i,j,k}^{2}}}{\beta_{l_{q}}}\geq\frac{\sqrt{\sum_{(i,j,k)\in{}G_{l_{q+1}}}Q({\bf f}^{(t)})_{i,j,k}^{2}}}{\beta_{l_{q+1}}}
\nonumber
\end{eqnarray}
holds for any $q\leq{}m-1$, and $m^{\star}=\min\{z:\mathrm{cardinality}(\cup_{q=1}^{z}G_{l_{q}})\geq{}\mathrm{cardinality}(g)\}$. The projection is given by ${\boldsymbol\theta}^{(t)}=Q^{-1}({\boldsymbol\vartheta}^{(t)})$ where
\begin{eqnarray}\label{eq:w-L21-norm-define}
{\boldsymbol\vartheta}^{(t)}_{i,j,k}&=&Q({\bf f}^{(t)})_{i,j,k}\max\left\{1-\frac{\beta_{l}\sqrt{\sum_{(i,j,k)\in{}G_{l_{m^{\star}+1}}}Q({\bf f}^{(t)})_{i,j,k}^{2}}}{\beta_{l_{m^{\star}+1}}\sqrt{\sum_{(i,j,k)\in{}G_{l}}Q({\bf f}^{(t)})_{i,j,k}^{2}}},0\right\},\quad\forall\,(i,j,k)\in{}G_{l}.
\nonumber
\end{eqnarray}
\end{enumerate}

\section{Results}

\begin{figure}[hbtp]
\centering
\includegraphics[width=5.25in]{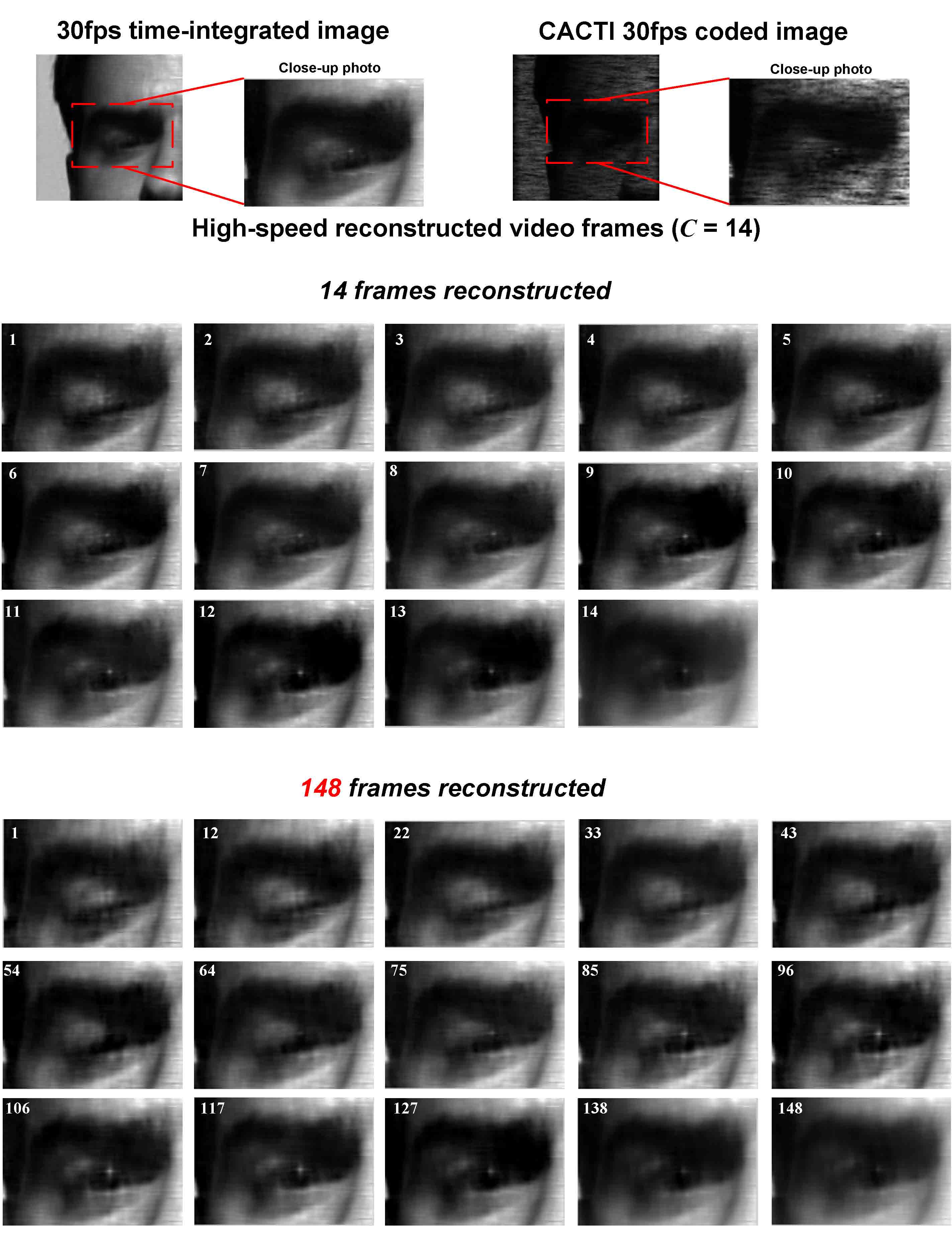}
\caption{High-speed ($C = 14$) video of an eye blink, from closed to open,
reconstructed from a single coded snapshot for $N_F = 14$ 
and $N_F = 148$.
The numbers on the bottom-right of the pictures represent the frame number of the video sequence.  Note that the eye is the only part of the scene that moves.  The top left frame shows the sum of these reconstructed frames, which approximates the motion captured by a 30fps camera without a coded aperture modulating the focal plane.}
\label{face}
\end{figure}
\begin{figure}[hbtp]
\centering
\includegraphics[width=5.25in]{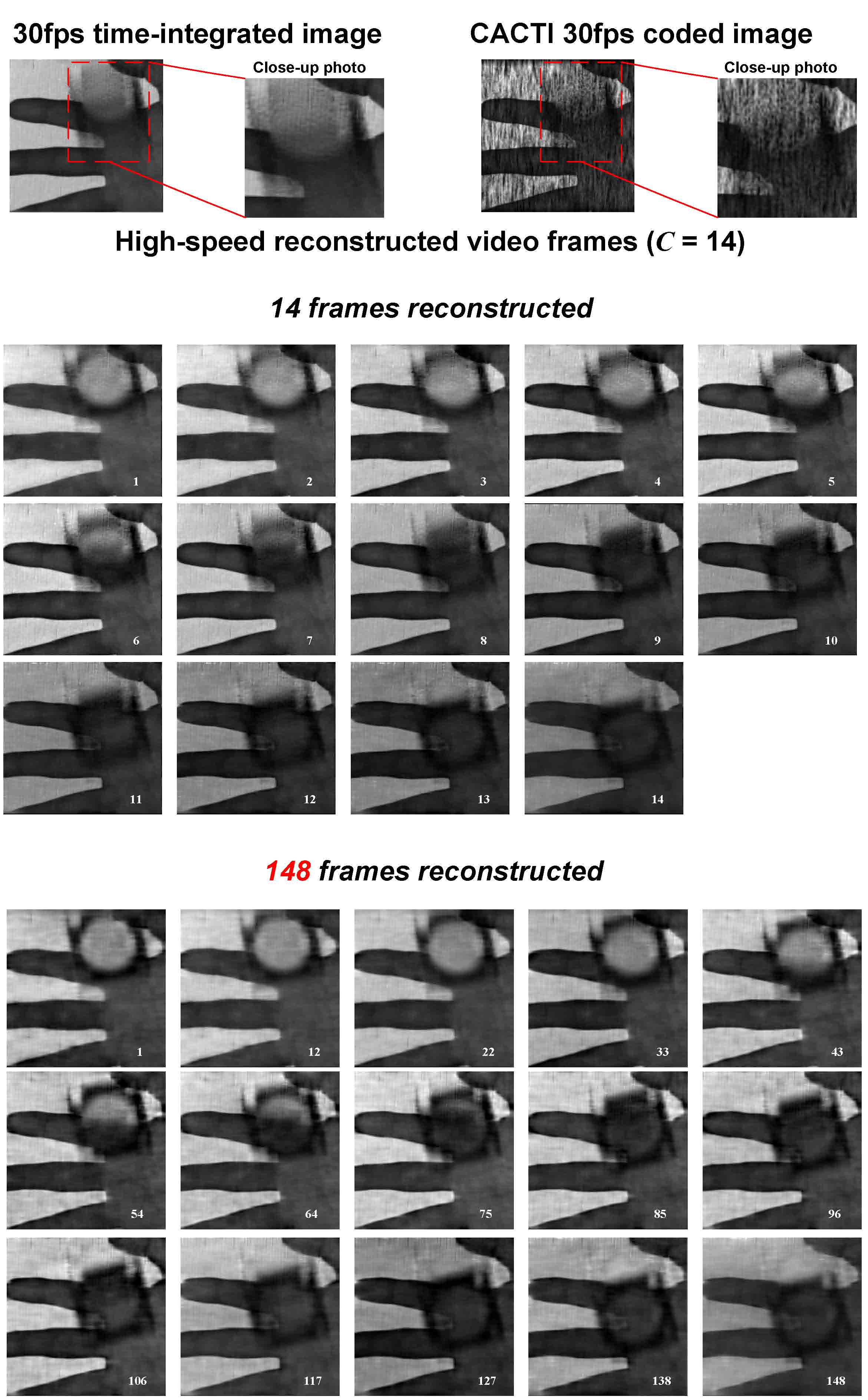}
\caption{Capture and reconstruction of a lens falling in front of a hand for $N_F = 14$ and $N_F = 148$.  Notice the reconstructed frames capture the magnification effects of the lens as it passes in front of the hand.}
\label{lens}
\end{figure}
\begin{figure}[hbtp]
\centering
\includegraphics[width=5.25in]{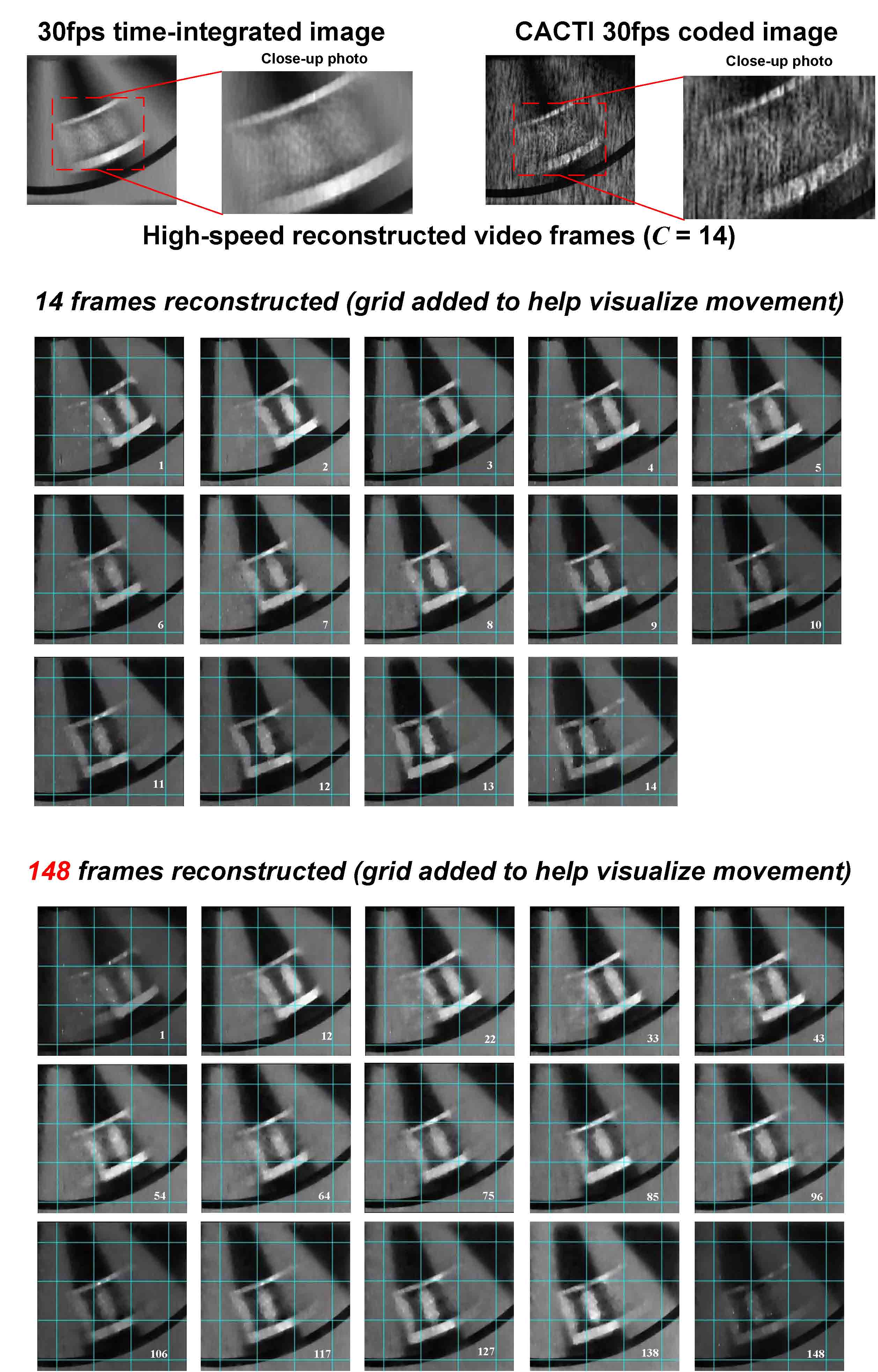}
\caption{Capture and reconstruction of a letter 'D' placed at the edge of a chopper wheel rotating at 15Hz for $N_F = 14$ and $N_F = 148$.  The white part of the letter exhibits ghosting effects in the reconstructions due to ambiguities in the solution.  The TwIST algorithm was used to reconstruct this data \cite{TWIST}.}
\label{fan}
\end{figure}
\begin{figure}[hbtp]
\centering
\includegraphics[width=5.25in]{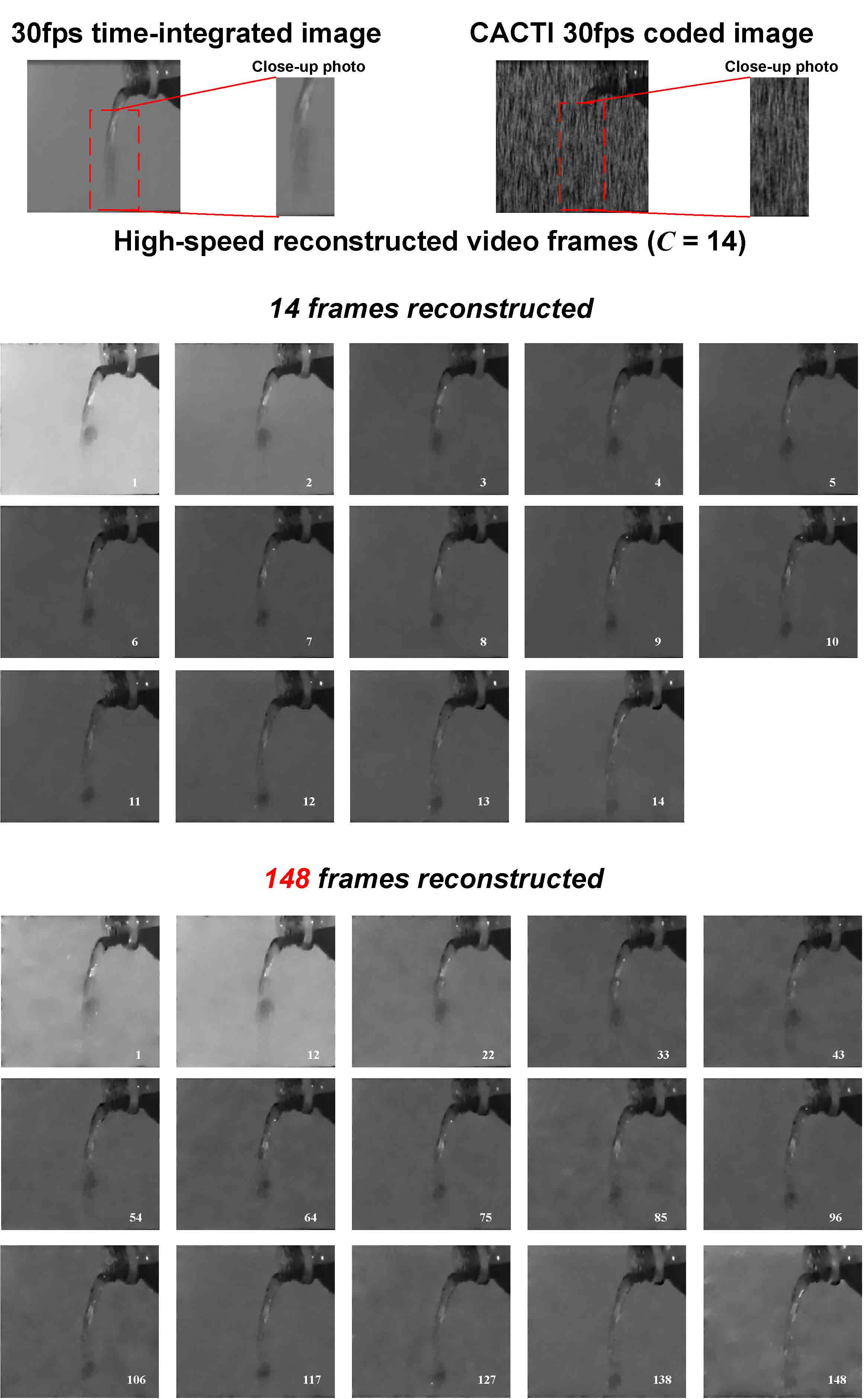}
\caption{Capture and reconstructed video of a bottle pouring water for $N_F = 14$ and $N_F = 148$.  Note the time-varying specularities in the video.  The TwIST algorithm was used to reconstruct this data \cite{TWIST}.}
\label{water}
\end{figure}

CACTI's experimental temporal superresolution results are shown in Figs. \ref{face}-\ref{water}.  An eye blinking, a lens falling in front of a hand, a chopper wheel with the letters `DUKE' placed on the blades, and a bottle pouring water into a cup are captured at 30fps and reconstructed with $N_F = C = 14$ and $N_F = 148$.  There is little aesthetic difference between these reconstructions.  In some cases, as with the lens and hand reconstruction, $N_F = 148$ appears to yield additional temporal information over $N_F = 14$.  The upper-left images depict the sum of the reconstructed frames, showing the expected time-integrated snapshots acquired with a 30fps video camera lacking spatiotemporal image plane modulation.

Note that several of these features, particularly the water pouring, are hardly visible among the moving code pattern.  Objects exhibiting large temporal motion blur were reconstructed with GAP using DCT bases, while wavelet bases were used for the stationary reconstructions.

The compression ratio $C$ is 14 rather than 16 because the triangle wave's peak and trough (${\bf T}{s_1}$ and ${\bf T}{s_{16}}$) are not accurately characterized by linear motion due to the mechanical deceleration time and were hence not placed into ${\bf H}$ to reduce model error.

\begin{figure}[hbtp]
\centering
\includegraphics[width=3.25in]{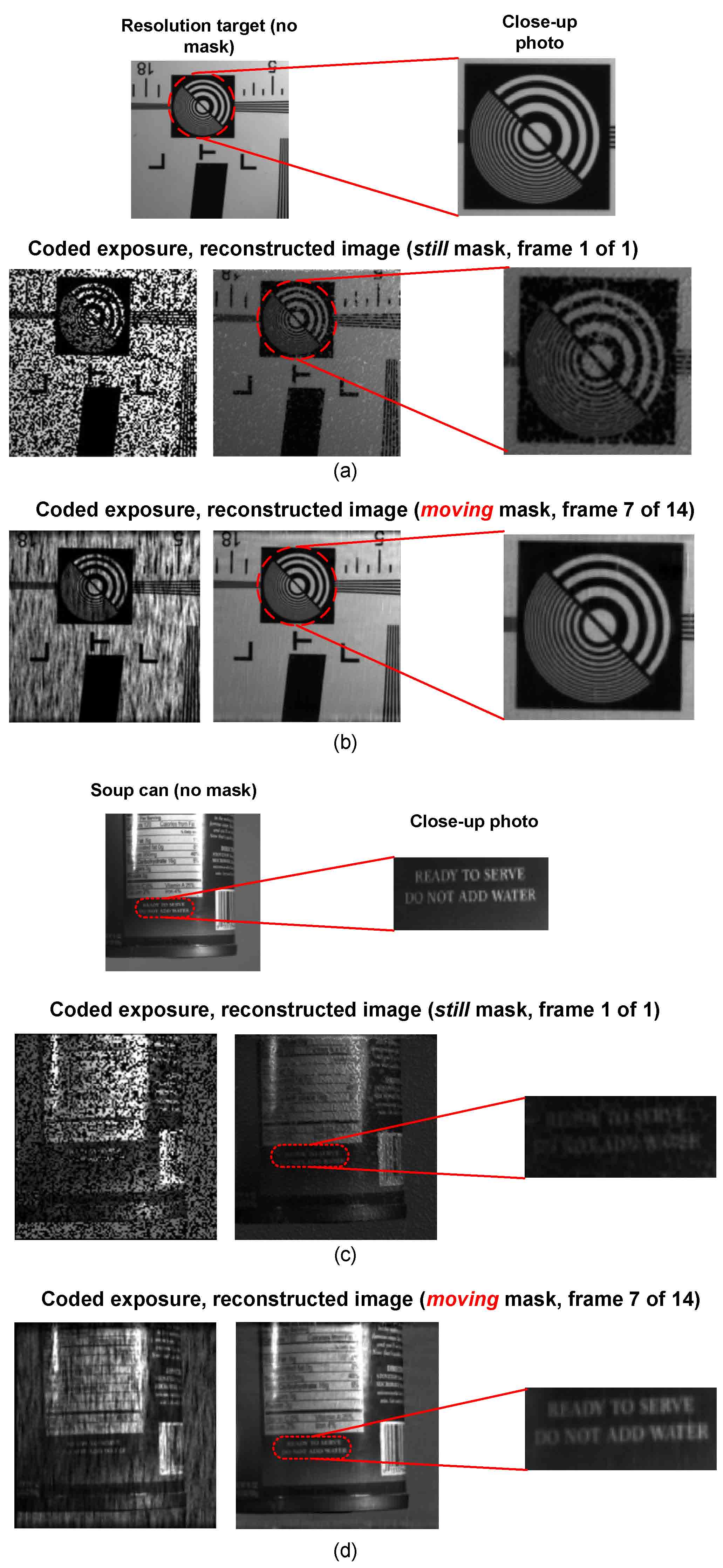}
\caption{Spatial resolution tests of (a,b) an ISO 12233 resolution target and (c,d) a soup can.  These objects were kept stationary several feet away from the camera.  (a,c) show reconstructed results without temporally moving the mask; (b,d) show the same objects when reconstructed with temporal mask motion.}
\label{spatialresult}
\end{figure}

\begin{figure}[hbtp]
\centering
\includegraphics[width=5.25in]{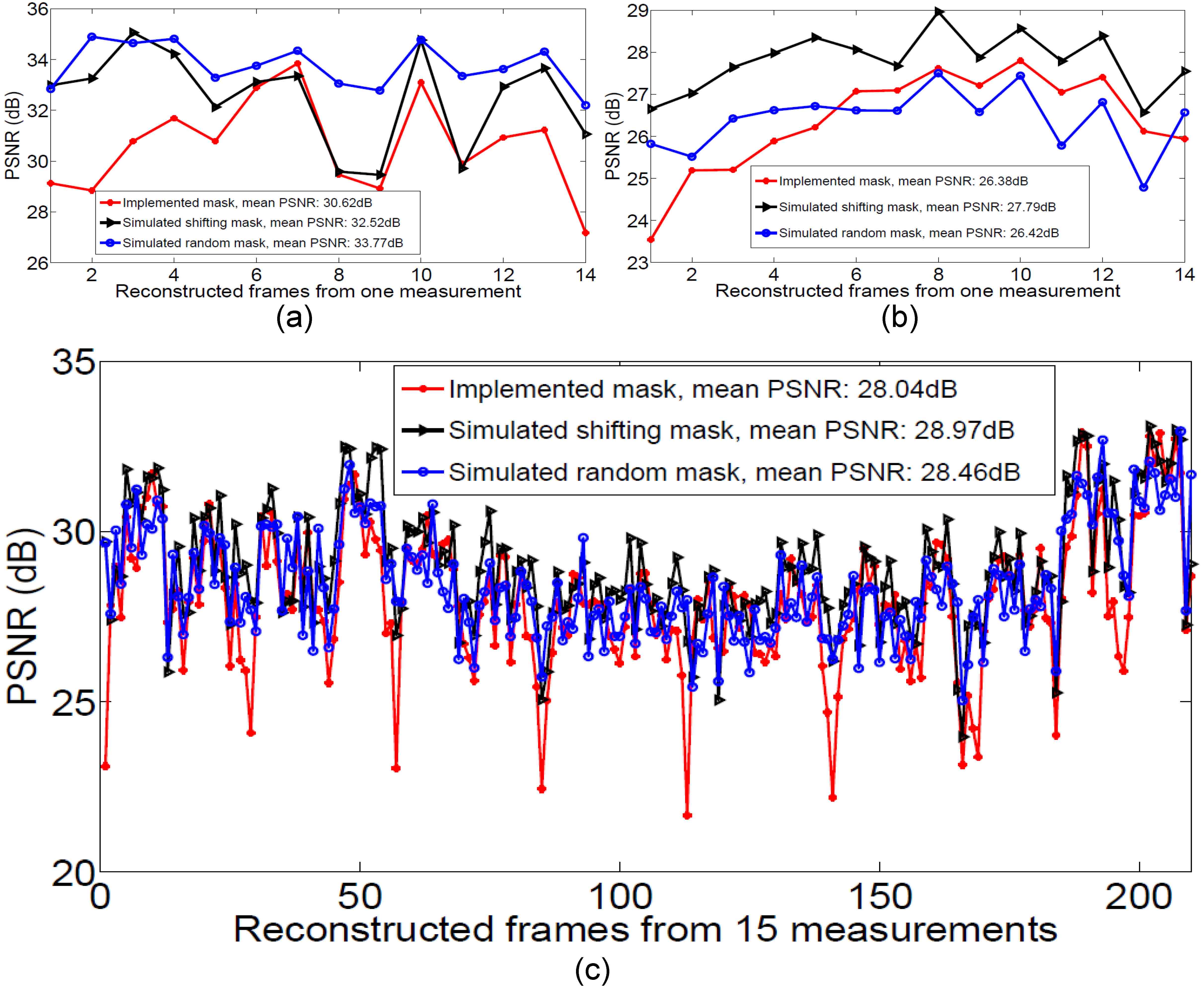}
\caption{Simulated and actual reconstruction PSNR by frame. (a), (b), and (c) show PSNR by high-speed, reconstructed video frame for 14 eye blink frames, 14 fan frames, and 210 fan frames, respectively, from snapshots ${\bf g}$.  Implemented mechanically-translated masks (red curves), simulated translating masks (black curves), and simulated LCoS coding (blue curves) were applied to high-speed ground truth data and reconstructed using GAP.}
\label{simulations}
\end{figure}

The CACTI system captures more unique coding projections of scene information when the mask moves $C$ pixels during the exposure (Fig. \ref{spatialresult}(b,d)) than if mask is held stationary (Fig. \ref{spatialresult}(a,c)), thereby improving the reconstruction quality for detailed scenes.  The stationary binary coded aperture may completely block small features, rendering the reconstruction difficult and artifact-ridden.

Completely changing the coding patterns $C$ times during $\Delta_t$ in hardware is only possible with adequate fill-factor employing a reflective LCoS device to address each pixel $C$ times during the integration period.  To compare the reconstruction fidelity of the low-bandwidth CACTI transmission function (Eq. (\ref{eqn:transmissionfunction})) with this modulation strategy, we present simulated PSNR values of videos in Fig. \ref{simulations}(a,b,c).  For these simulations, reconstructed experimental frames at $N_F = 14$ were summed to emulate a time-integrated image.  The high-speed reconstructed frames were used as ground truth.  We reapply (1) the actual mask pattern; (2) a simulated CACTI mask moving with motion $s_k$; and (3) a simulated, re-randomized coding pattern to each high-speed frame used as ground truth.  The reconstruction performance difference between translating the same code and re-randomized the code for each of the $N_F$ reconstructed frames is typically within 1dB.

\section{Discussion and Conclusion}
\label{Sec:Conclusion}

CACTI presents a new framework to uniquely code and decompress high-speed video exploiting conventional sensors with limited bandwidth.  This approach benefits from mechanical simplicity, large compression ratios, inexpensive scalability, and extensibility to other frameworks.

We have demonstrated GAP, a new reconstruction algorithm that can use one of several bases to compactly represent a sparse signal.  This fast-converging algorithm requires no prior knowledge of the target scene and will scale easily to lager image sizes and compression ratios.  This algorithm was used for all reconstructions except Figs. \ref{fan} and \ref{water}.

Despite GAP's computational efficiency, large-scale CACTI implementations will seek to minimize the data {\em reconstructed} in addition to that transmitted to sufficiently represent the optical datastream.  Future work will adapt the compression ratio $C$ such that the resulting reconstructed video requires the fewest number of computations to depict the motion of the scene with high quality. 


Since coded apertures are passive elements, extending the CACTI framework onto larger values of $N$ only requires use of a larger mask and a greater detector sensing area, making it a viable choice for large-scale compressive video implementations.  As $N$ increases, LCoS-driven temporal compression strategies must modulate $N$ pixels $C$ times per integration.  Conversely, translating a passive transmissive element attains $C$ times temporal resolution without utilizing any additional bandwidth relative to conventional low-framerate capture.  

Future large-scale imaging systems may employ CACTI's inexpensive coding strategy in conjunction with higher-dimensional imaging modalities, including spectral compressive video.  Integrating CACTI with the current CASSI system \cite{gehm2007single} should provide preliminary reconstructions depicting 4-dimensional datasets ${\bf f}(x,y,\lambda,t)$.

\end{document}